\documentclass[runningheads]{llncs}

\usepackage{hyperref}
\usepackage{graphics}
\usepackage{graphicx}
\usepackage{arabtex}
\usepackage{utf8}
\usepackage{xcolor}

\usepackage{xcolor}
\usepackage{examples}
\usepackage{xargs}    
\usepackage{comment}

\usepackage[colorinlistoftodos,prependcaption,textsize=tiny]{todonotes}
\newcommandx{\fbcom}[2][1=]{\todo[#1]{#2}}
\newcommandx{\bgcom}[2][1=]{\todo[linecolor=blue,backgroundcolor=blue!25,bordercolor=blue,#1]{#2}}
\newcommandx{\jkcom}[2][1=]{\todo[linecolor=magenta,backgroundcolor=magenta!25,bordercolor=magenta,#1]{#2}}
\newcommandx{\vmcom}[2][1=]{\todo[linecolor=yellow,backgroundcolor=yellow!25,bordercolor=yellow,#1]{#2}}

\begin{document}
\setcode{utf8}
\title{Irony Detection in a Multilingual Context}
\titlerunning{Irony Detection in a Multilingual Context}

\author{
Bilal Ghanem \inst{1} \and
Jihen Karoui\inst{2} \and
Farah Benamara\inst{3} \and
Paolo Rosso \inst{1} \and
V\'eronique Moriceau\inst{3}
}
\authorrunning{Ghanem et al.}

\institute{
PRHLT Research Center, Universitat Polit\`ecnica de Val\`encia, Spain \\
\email{\{bigha@doctor, prosso@dsic\}.upv.es} \and
AUSY R\&D, Paris, France \\
\email{jkaroui@ausy.fr} \and
IRIT, CNRS, Universit\'e de Toulouse, France \\
\email{\{benamara,moriceau\}@irit.fr}
}
%
\maketitle              
\begin{abstract}
This paper proposes the first multilingual (French,   English  and  Arabic) and multicultural   (Indo-European  languages vs.  less culturally close languages) irony detection system. 
We employ both feature-based models and neural architectures using monolingual word representation. We compare the performance of these systems with state-of-the-art systems to identify their capabilities. We show that these monolingual models trained separately on different languages using multilingual word representation or text-based features can open the door to irony detection in languages that lack of annotated data for irony.


\end{abstract}

\section{Motivations}
 Figurative language makes use of figures of speech to convey non-literal meaning~\cite{grice:1975,attardo:2000}. It encompasses a variety of phenomena, including metaphor, humor, and irony. We focus here on irony and uses it as an umbrella term that covers satire, parody and sarcasm.
 
Irony detection (ID) has gained relevance recently, due to its importance to extract information from texts. For example, to go beyond the literal matches of user queries, Veale enriched information retrieval with new  operators to enable the non-literal retrieval of creative expressions \cite{Veale:2011}. Also,  
the performances of sentiment analysis systems  drastically decrease when applied to ironic texts~\cite{Deft2017,farias2016irony}. 
Most related work concern English \cite{SemEval:2018,huang2017irony} with some
efforts 
in French \cite{Karoui:2015}, Portuguese \cite{carvalho:2009}, Italian \cite{gianti:2012}, Dutch \cite{liebrecht:2013}, Hindi \cite{Swami:18}, Spanish variants \cite{ortega2019overview} and Arabic \cite{KarouiACLING:2017,idat2019}. 
Bilingual ID with one model per language has also been explored, like English-Czech \cite{Techc:2014} and English-Chinese \cite{Tang}, but not within a cross-lingual perspective.

In social media, such as Twitter, specific hashtags (\#irony, \#sarcasm) are often used as gold labels to detect irony in a supervised learning setting.  Although  recent studies pointed out the  issue  of  false-alarm  hashtags in self-labeled data~\cite{Huang:2018}, ID via hashtag  filtering provides researchers positive examples with high precision. On the other hand, 
systems are not able to detect irony in  languages where such filtering is not always possible.
Multilingual prediction (either relying on machine translation  or multilingual embedding methods) is a common solution to tackle under-resourced languages 
\cite{Bikel:2012,Ruder17}.
While  multilinguality has been widely investigated in  information retrieval \cite{Litschko:2018,sasaki-etal-2018-cross} and several NLP tasks (e.g.,    sentiment analysis  \cite{Balahur:2014,Barnes:2018} 
and named entity recognition \cite{Jian:2017}), 
no one explored it for irony. 

We aim here to bridge the gap by tackling ID in tweets from both  multilingual (French, English and Arabic) and multicultural perspectives (Indo-European languages  whose  speakers  share  quite  the same   cultural   background vs.  less culturally close languages). Our approach does not rely either on machine translation or parallel corpora (which are not always available), but rather builds on previous corpus-based studies  that show that irony is a universal phenomenon and many languages share similar irony devices. For example, 
Karoui et. al~\cite{Karoui:2017} concluded that their multi-layer annotated schema, initially used to annotate French tweets, is portable  to  English  and  Italian,  observing  relatively  the  same tendencies in terms of irony categories and markers. 
Similarly, Chakhachiro~\cite{chakhachiro:2007} studies irony in English and Arabic, and  shows that both languages share several similarities in the rhetorical (e.g., 
overstatement), grammatical (e.g., redundancy
) and lexical (e.g., synonymy) usage of irony devices. 
The next step now is to show to what extent these observations are still valid from  a computational point of view. 
Our contributions are:
\renewcommand{\theenumi}{\Roman{enumi}}%
\begin{enumerate}
   \item \textit{A new freely available corpus of Arabic tweets} manually annotated for irony detection\footnote{The corpus is available at \url{https://github.com/bilalghanem/multilingual_irony}}.  
    \item \textit{Monolingual ID}: We propose both feature-based models (relying
on  language-dependent and language-independent features) and  neural models to
measure to what extent ID is language dependent. 
    \item \textit{Cross-lingual  ID}: 
       We experiment using cross-lingual word representation by training on one language and testing on another one to measure how the proposed models are culture-dependent. Our results are encouraging and open the door to ID in languages that lack of annotated data for irony.
\end{enumerate}


\section{Data}

 \textbf{Arabic dataset} (\textsc{Ar}=$11,225$ tweets). Our starting point was the corpus built by \cite{KarouiACLING:2017} 
 that we extended to different political issues and events related to the Middle East and Maghreb that hold during the years $2011$ to $2018$.  Tweets were collected using a set of predefined keywords (which targeted specific political figures or events) and containing or not Arabic ironic hashtags (\<سخرية>\#, \<مسخرة>\#, \<تهكم>\#, \<استهزاء>\#) \footnote{All of these words are synonyms where they mean "Irony".}.
 The collection process resulted in a set of 
 $6,809$ ironic tweets ($I$) vs. $15,509$ non ironic ($NI$) written using standard (formal) and different Arabic language varieties: Egypt, Gulf, Levantine, and Maghrebi dialects.
 
To investigate the validity of using the original tweets labels, a  sample of $3,000$ $I$ and $3,000$ $NI$ was manually annotated by two Arabic native speakers which resulted in $2,636$ $I$ vs. $2,876$ $NI$. The inter-annotator agreement using Cohen's Kappa was $0.76$, while the agreement score between the annotators'~labels and the original labels was $0.6$.
Agreements being relatively good knowing the difficulty of the task, we sampled $5,713$ instances from the original unlabeled dataset to our manually labeled part. The added tweets have been manually checked  
to remove duplicates, very short tweets and tweets that depend on external links, images or videos to understand their meaning.

\textbf{French dataset} (\textsc{Fr}=$7,307$ tweets). We rely on the corpus used for the DEFT 2017 French shared task on irony \cite{Deft2017} which consists of  tweets relative to a set of topics discussed in the media between  2014 and 2016 and contains topic keywords and/or French irony hashtags (\#ironie, \#sarcasme). Tweets have been annotated by three annotators (after removing the original labels) with a reported Cohen's Kappa of $0.69$.

\textbf{English dataset} (\textsc{En}=$11,225$ tweets). We use the corpus built by \cite{Techc:2014} which consists of $100,000$ tweets collected  using the hashtag \#sarcasm. 
It was used as benchmark in several works \cite{ghanem2018ldr,farias2017sentiment}. We sliced a subset of approximately $11,200$ tweets to match the sizes of the other languages' datasets.

Table \ref{data} shows the tweet distribution in all corpora. 
Across the three languages, we keep a similar number of instances for train and test sets 
to have fair cross-lingual experiments as well (see Section \ref{CL}). Also, for French, we use the original dataset without any modification, keeping the same number of records for train and test to better compare with state-of-the-art results. 
For the classes distribution (ironic vs. non ironic), we do not choose a specific ratio but we use the resulted distribution from the random shuffling process.

\vspace{-0.5cm}

\begin{footnotesize}
\begin{table}[!h]
\centering
\caption{Tweet distribution in all corpora.}
\begin{tabular}{|c|c|c||c|c|}
  \hline
 & \textbf{\# Ironic} & \textbf{\# Not-Ironic} & \textbf{Train}&\textbf{Test}\\
  \hline
  \textsc{Ar}   
    & $6,005$ & $5,220$ & $10,219$& $1,006$ \\
    \hline
  \textsc{Fr}  &    $2,425$ & $4,882$ & $5,843$& $1,464$   \\ 
    \hline
    \textsc{En}     & $5,602$ & $5,623$ & $10,219$& $1,006$  \\
  \hline
\end{tabular}
\label{data} 
\end{table}

\end{footnotesize}

\vspace{-0.8cm}

\section{Monolingual Irony Detection}
It is important to note that our aim is not to outperform state-of-the-art models in monolingual ID but to investigate which of the monolingual architectures (neural or feature-based) can achieve comparable results with existing  systems. The result can show which kind of features works better in the monolingual settings and can be employed to detect irony in a multilingual setting. In addition, it can show us to what extend ID is language dependent by comparing their results to multilingual results. 
Two models have been built, as explained below. Prior to learning, basic preprocessing steps were performed for each language (e.g., removing foreign characters, ironic hashtags, mentions, and URLs). 

\textbf{Feature-based models.} We used state-of-the-art features that have shown to be useful in ID: some of them are language-independent (e.g., punctuation marks, positive and negative emoticons, quotations, personal pronouns, tweet's length, named entities) while others are language-dependent relying on dedicated lexicons (e.g., negation, opinion lexicons, opposition words). Several classical machine learning classifiers were tested with several feature combinations, among them Random Forest (RF) achieved the best result with all features. 

\textbf{Neural model with monolingual embeddings.} We used Convolutional Neural Network (CNN) network whose structure is similar to the one proposed by \cite{Kim:2014}. 
For the embeddings, we relied on $AraVec$ \cite{Soliman:2017} for Arabic,  FastText \cite{Grave:2018} for French, and Word2vec Google News  \cite{mikolov-etal-2013-linguistic} for English \footnote{
Other available pretrained embeddings models have also been tested. }. 
For the three languages, the size of the embeddings is $300$ and the embeddings were fine-tuned during the training process. The CNN network was tuned with 20\% of the training corpus using the $Hyperopt$\footnote{\url{https://github.com/hyperopt/hyperopt}} library. 



\textbf{Results.} Table \ref{ResMonolingual} shows the results obtained when using  train-test configurations for each language. 
For English,
our results, in terms of macro F-score ($F$), were not comparable to those of \cite{Techc:2014,tay-etal-2018-reasoning}, 
as we used 11\% of the original dataset. For French, our scores  are in line with those reported in state of the art (cf. best system in the irony shared task  achieved  $F=78.3$ \cite{Deft2017}). They outperform those obtained for Arabic ($A=71.7$) \cite{KarouiACLING:2017} and are comparable to those recently reported in the irony detection shared task in Arabic tweets \cite{idat2019,ghanem2019idat} ($F=84.4$). Overall, the results show that semantic-based information captured by the embedding space are more productive comparing to standard surface and lexicon-based features. 

\vspace{-0.2cm}

\begin{footnotesize}
\begin{table}[ht]
    \centering
    \caption{Results of the monolingual experiments (in percentage) in terms of accuracy (A), precision (P), recall (R), and macro F-score (F).} 
    \begin{tabular}{|c|c|c|c|c||c|c|c|c||c|c|c|c|}
        \hline
        \multicolumn{5}{|c||}{\bf \textsc{Arabic}}
        & \multicolumn{4}{c||}{\bf \textsc{French}}
        & \multicolumn{4}{c|} {\bf \textsc{English}}\\
        \hline
         & A & P & R & F & A & P & R& F& A &  P & R& F\\
        \hline
        RF & 
        68.0 & 67.0 & 82.0 & 68.0 & 
        68.5 & 71.7 & 87.3 & 61.0 &
        61.2 & 60.0 & 70.0 & 61.0 
        \\
        CNN &
        \textbf{80.5} & 79.1 & 84.9 & \textbf{80.4} & 
        \textbf{77.6} & 68.2 & 59.6 & \textbf{73.5} & 
        \textbf{77.9} & 74.6 & 84.7 & \textbf{77.8} \\
        \hline
    \end{tabular}
    \label{ResMonolingual} 
\end{table}
\end{footnotesize}

\vspace{-0.9cm}

\section{Cross-lingual Irony Detection}
\label{CL}
We use the previous CNN architecture with bilingual embedding and the RF model with surface features (e.g., use of personal pronoun, presence of interjections, emoticon or specific punctuation)\footnote{To avoid language dependencies, we rely on surface features only discarding those that require external semantic resources or morpho-syntactic parsing.} to verify which pair of the three languages: (a) has similar ironic pragmatic devices, and (b) uses similar text-based pattern in the narrative of the ironic tweets.
As continuous word embedding spaces exhibit similar structures across (even distant) languages \cite{Mikolov:2013}, we use a multilingual word representation which aims to learn a linear mapping from a source to a target embedding space. 
Many methods have been proposed to learn this mapping such as parallel data supervision  and bilingual dictionaries \cite{Mikolov:2013}  or unsupervised methods relying on monolingual corpora \cite{Conneau:2017,Artetxe:2018,Wada:2018}. For our experiments, we use Conneau et al 's approach as it showed superior results with respect to the literature  \cite{Conneau:2017}. 
We perform several experiments by training on one language ($lang_1$) and testing on another one ($lang_2$) (henceforth $lang_1\rightarrow lang_2$). We get 6 configurations, 
plus two others to evaluate how irony devices are expressed cross-culturally, i.e. in European vs. non European languages. In each experiment, we took 20\% from the training 
to validate the model before the testing process. Table \ref{ResMultilingual} presents the results. 

\vspace{-0.8cm}

\begin{footnotesize}
\begin{table}[ht]
    \centering
    \caption{Results of the cross-lingual experiments. }
    \begin{tabular}{|c||c|c|c|c||c|c|c|c|}
        \hline
      {}   & \multicolumn{4}{c||} {CNN} & \multicolumn{4}{c|} {RF} \\
        \hline
      Train$\rightarrow$Test   & A  & P & R & F & A  & P & R & F \\
        \hline
Ar$\rightarrow$Fr & 60.1 & 37.2 & 26.6 & \textbf{51.7} & 47.03 & 29.9 & 43.9 & 46.0 \\
Fr$\rightarrow$Ar & 57.8 & 62.9 & 45.7 & \textbf{57.3} & 51.11 & 61.1 & 24.0 & 54.0 \\ 
 \hline 
Ar$\rightarrow$En & 48.5 & 26.5 & 17.9 & 34.1 & 49.67 & 49.7 & 66.2 & \textbf{50.0}  \\
En$\rightarrow$Ar& 56.7 & 57.7 & 62.3 & \textbf{56.4} & 52.5 & 58.6 & 38.5 & 53.0  \\ 
 \hline 
Fr$\rightarrow$En & 53.0 & 67.9 & 11.0 & 42.9 & 52.38 & 52.0 & 63.6 & \textbf{52.0}  \\
En$\rightarrow$Fr  & 56.7 & 33.5 & 29.5 & 50.0 & 56.44 & 74.6 & 52.7 & \textbf{58.0}  \\ 
 \hline
(En/Fr)$\rightarrow$Ar & 62.4 & 66.1 & 56.8 & \textbf{62.4} & 55.08 & 56.7 & 68.5 & 62.0 \\
  Ar$\rightarrow$(En/Fr) & 56.3 & 33.9 & 09.5 & 42.7 & 59.84 & 60.0 & 98.7 & \textbf{74.6}  \\ 
  \hline
    \end{tabular}
    \label{ResMultilingual} 
\end{table}
\end{footnotesize}

\vspace{-1.1cm}

From a semantic perspective, 
despite the language and cultural differences between Arabic and French languages, CNN results show a high performance comparing to the other languages pairs when we train on each of these two languages and test on the other one. Similarly, for the French and English pair, but when we train on 
French 
they are quite lower. We have a similar case when we train on Arabic and test on English. We can justify that by, the language presentation of the Arabic and French tweets are quite informal and have many dialect words that may not exist in the pretrained embeddings we used comparing to the English ones (lower embeddings coverage ratio), which become harder for the CNN to learn a clear semantic pattern. Another point is the presence of Arabic dialects, where some dialect words may not exist in the multilingual pretrained embedding model that we used. On the other hand, from the text-based perspective, 
the results show that the text-based features can help in the case when the semantic aspect shows weak detection; this is the case for the $Ar\longrightarrow En$ configuration. It is worthy to mention that the highest result we get in this experiment is from the En$\rightarrow$Fr pair, as 
both languages use Latin characters. 
Finally, when investigating the relatedness between European vs. non European languages (cf. (En/Fr)$\rightarrow$Ar), we 
obtain similar results than those obtained in the monolingual experiment (macro F-score 62.4 vs. 68.0) and best results are achieved by  Ar $\rightarrow$(En/Fr). This shows that there are pragmatic devices in common between both sides and, in a similar way, similar text-based patterns in the narrative way of the ironic tweets.


\section{Discussions and Conclusion}
This paper proposes the first multilingual 
ID in tweets. We show that simple monolingual architectures (either neural or feature-based) trained separately on each language can be successfully used in a multilingual setting providing a cross-lingual word representation or basic surface features. 
Our monolingual results are comparable to state of the art for the three languages. 
The CNN architecture trained on cross-lingual word representation shows that irony has a certain similarity between the languages we targeted despite the cultural differences which confirm that irony is a universal phenomena, as already shown in previous linguistic studies \cite{sigar:2012,Karoui:2017,colston2019irony}. 
The manual analysis of the common misclassified tweets across the languages in the multilingual setup, shows that classification errors are due to three 
main factors. (1) First, the \textit{absence of context}  where writers did not provide sufficient information to capture the ironic sense even in the monolingual setting, as in \< نبدا تاني يسقط يسقط حسني مبارك !! > (\textit{Let's start again, get off get off Mubarak!!}) where the writer mocks the Egyptian revolution, as the actual president "Sisi" is viewed as Mubarak's fellows.
 (2) Second, the presence of \textit{out of vocabulary (OOV) terms} because of the weak coverage of the mutlilingual embeddings which make the system fails to generalize when the OOV set of unseen words is large during the training process. We found tweets in all the three languages written in a very informal way, where some characters of the words were deleted, duplicated or written phonetically (e.g \textit{phat} instead of \textit{fat}). (3) Another important issue is the difficulty to \textit{deal with the Arabic language}. Arabic tweets are often characterized by non-diacritised texts, a large variations of unstandardized dialectal
Arabic (recall that our dataset has 4 main varieties, namely Egypt, Gulf, Levantine, and Maghrebi),  presence of transliterated words (e.g. the word \textit{table} becomes \<طابلة> (\textit{tabla})), and finally linguistic code switching between Modern Standard Arabic and several dialects,
and between Arabic and other languages like English and French. We found some tweets contain only words from one of the varieties and most of these words do not exist in the Arabic embeddings model. For example in  \<مبارك بقاله كام يوم مامتش .. هو عيان ولاه ايه \#مصر  >
   (\textit{Since many days Mubarak didn't die .. is he sick or what? \#Egypt}), only the words \<يوم> (day), \<مبارك> (Mubarak), and \<هو> (he) exist in the embeddings. Clearly, considering only these three available words, we are not able to understand the context or the ironic meaning of the tweet. 
   
To conclude, our multilingual experiments confirmed that the door is open towards multilingual approaches for ID. 
Furthermore, our results showed that ID can be applied to languages that lack of annotated data. Our next step is to experiment with other languages such as Hindi and Italian.

\section*{Acknowledgment}
The work of Paolo Rosso was partially funded by the Spanish MICINN under the research project MISMIS-FAKEnHATE (PGC2018-096212-B-C31).

%
%
%
\bibliographystyle{splncs04}
\bibliography{references}
\end{document}